\documentclass{article}

\PassOptionsToPackage{sort&compress,numbers,square,comma}{natbib}
\usepackage[preprint]{arxiv_lrcp}

\usepackage[utf8]{inputenc}
\usepackage[T1]{fontenc}
\usepackage[colorlinks=true,linkcolor=black,citecolor=black,urlcolor=blue]{hyperref}
\usepackage{url}
\usepackage{booktabs}
\usepackage{amsmath,amssymb,amsfonts,amsthm,mathtools,bm}
\usepackage{microtype}
\usepackage[table]{xcolor}
\usepackage{array}
\usepackage{graphicx}
\usepackage{subcaption}
\usepackage{algorithm}
\usepackage{algpseudocode}
\usepackage{enumitem}
\usepackage{multirow}
\setcitestyle{numbers,square,comma,sort&compress}

\makeatletter
\renewcommand{\@notice}{%
  \enlargethispage{2\baselineskip}%
  \@float{noticebox}[b]%
    \footnotesize\textsuperscript{*}Corresponding authors.\quad\textsuperscript{\ensuremath{\ddagger}}Project leader.%
  \end@float%
}
\makeatother

\newcommand{\ours}{\textbf{LRCP}}
\newcommand{\best}[1]{\textbf{#1}}
\newcommand{\second}[1]{\underline{#1}}
\definecolor{flowgray}{RGB}{238,238,238}
\definecolor{flowgreenbg}{RGB}{231,242,225}
\definecolor{flowgreen}{RGB}{34,139,34}
\definecolor{flowtextgray}{RGB}{122,122,122}
\newcommand{\upperboundrow}[2]{%
  \rowcolor{flowgray}
  \multicolumn{#1}{c}{\textit{Upper Bound, #2} \textbf{(100\%)}}\\
}
\newcommand{\retainrow}[3]{%
  \rowcolor{flowgray}
  \multicolumn{#1}{c}{\textit{Retain #2 Tokens} {\color{flowgreen}($\downarrow$ #3)}}\\
}
\newcommand{\retainlabelrow}[3]{%
  \rowcolor{flowgray}
  \multicolumn{#1}{c}{\textit{#2} {\color{flowgreen}($\downarrow$ #3)}}\\
}

\title{LRCP: Low-Rank Compressibility Guided Visual Token Pruning for Efficient LVLMs}

\author{%
  \textbf{Hongyu Lu}$^{1,2}$ \quad
  \textbf{Feng Zhang}$^{1,\ddagger}$ \quad
  \textbf{Wenwei Jin}$^{1}$ \quad
  \textbf{Huanling Hu}$^{3}$ \\
  \textbf{Tianjun Shi}$^{2}$ \quad
  \textbf{Shikai Jiang}$^{2,*}$ \quad
  \textbf{Yao Hu}$^{1}$ \quad
  \textbf{Jiawei Li}$^{1,*}$ \\
  $^1$Xiaohongshu \quad
  $^2$Harbin Institute of Technology \quad
  $^3$Fudan University \\
  \texttt{24S021013@stu.hit.edu.cn, shitianjun@stu.hit.edu.cn} \\
  \texttt{\{zhangfeng4, wangdesheng\}@xiaohongshu.com} \\
  \texttt{hlhu24@m.fudan.edu.cn, wenwei1217.jin@gmail.com} \\
  \texttt{yaoohu@gmail.com, jiangshikai@hit.edu.cn}
}

\begin{document}

\maketitle

\begin{abstract}
Large vision-language models (LVLMs) achieve strong multimodal understanding, but their inference cost grows rapidly with the number of visual tokens, especially for high-resolution images and long videos.
Existing attention-based methods estimate token importance from attention scores, which may introduce positional bias, while representation-based methods reduce visual redundancy based on feature relations or reconstruction errors, overlooking the global structure of the visual token set.
In this paper, we revisit visual token compression from the perspective of low-rank compressibility.
Across models and datasets, we observe that visual token representations exhibit a pronounced low-rank structure, with a dominant subspace that remains stable even after a large fraction of tokens is randomly removed.
Motivated by this finding, we propose LRCP, a training-free compression framework that first estimates the dominant low-rank subspace of visual tokens via PCA, and then scores each token by its projection residual onto this subspace, retaining tokens that are poorly explained by the low-rank background.
Extensive experiments show that LRCP achieves superior results, preserving 94.7\% of the original image-understanding performance with an 88.9\% token reduction and 97.8\% of the average video-understanding accuracy with an 87.5\% token reduction.
\end{abstract}

\section{Introduction}

Large vision-language models (LVLMs) have substantially advanced multimodal understanding \citep{alayrac2022flamingo,li2023blip2,liu2024visual,dai2023instructblip,zhu2023minigpt4,openai2023gpt4v}, with strong results in visual question answering \citep{antol2015vqa,goyal2017vqa,singh2019textvqa,mathew2021docvqa}, image captioning \citep{lin2014coco,young2014flickr30k,agrawal2019nocaps}, and video understanding \citep{li2024llavaonevision,lin2023videollava,zhang2025videollama3}. As these models are applied to higher-resolution images \citep{chen2024internvl,li2024llavaonevision,guo2025seed15vl} and long videos \citep{lin2023videollava,zhang2025videollama3}, the number of visual tokens increases rapidly. 
This growth raises both computation and memory costs during prefilling and also increases inference latency. Reducing visual token overhead has therefore become critical for efficient LVLM deployment \citep{liu2025shifting}.

To address this, recent work has explored visual token compression, with existing methods broadly categorized as attention-based \citep{chen2024fastv,arif2025hired,zhang2024fastervlm,liu2025hiprune,yang2024visionzip} or representation-based \citep{zhang2024sparsevlm,xing2024pyramiddrop,zhang2025vscan,ma2026apet,shang2024prumerge,ye2025atpllava,wang2025folder,xing2025cvc}. Attention-based methods estimate token importance from attention scores, but they can be affected by positional bias and are less compatible with efficient attention implementations such as FlashAttention \citep{dao2022flashattention,dao2024flashattention2}. 
Representation-based methods reduce visual redundancy by exploiting relations among token features, but they mainly capture local relations and often rely on heuristic distance measures. As a result, existing methods do not explicitly model the global structure of the visual token set, which limits their ability to preserve important semantic information under high compression ratios.

This paper studies visual token compression from the perspective of low-rank compressibility. We observe that visual tokens in LVLMs concentrate around a small number of dominant directions in the feature space, forming an approximate low-rank structure. This suggests that, beyond individual token importance or local similarity, visual token redundancy is rooted in the global structure of visual representations. Specifically, this global structure can be described by a shared low-rank background. We further find that this dominant subspace remains highly stable even when a large fraction of tokens is randomly removed, indicating that the low-rank structure is a collective property rather than driven by a few specific tokens. Based on these observations, we formulate visual token compression as a lossy compression problem under a low-rank background, where tokens with larger deviations from the dominant subspace are more likely to carry discriminative visual information.

\begin{figure*}[t!]
\centering
\includegraphics[width=\textwidth]{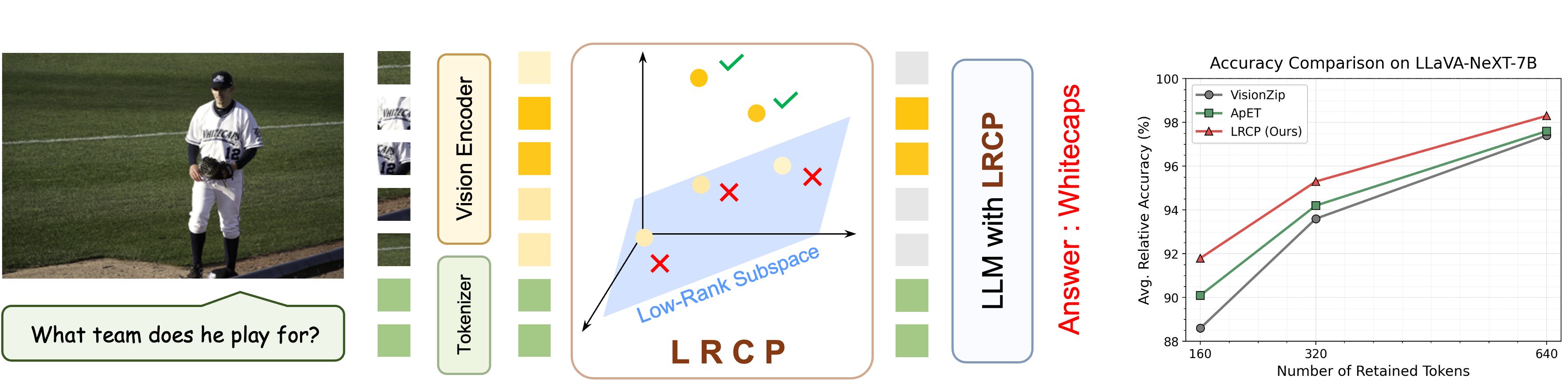}
\caption{\textbf{(Left)} LRCP estimates the dominant low-rank subspace of visual tokens via PCA and selects tokens based on their projection residuals. \textbf{(Right)} Accuracy comparison with LLaVA-NeXT-7B across 7 benchmarks, showing that ours outperforms both VisionZip (attention-based) and ApET (representation-based).}
\label{fig:low_rank_visualization}
\end{figure*}

Motivated by these observations, we propose LRCP (Low-Rank Compressibility Guided Pruning), a training-free visual token compression framework. LRCP estimates the dominant low-rank subspace via PCA and scores each token by its projection residual (Figure~\ref{fig:low_rank_visualization}, left). Under the same compression ratio, LRCP outperforms both attention-based and representation-based methods (Figure~\ref{fig:low_rank_visualization}, right). The method can be inserted after the visual encoder output and at intermediate LLM layers. Since it does not require attention weights, it is fully compatible with efficient inference implementations \citep{dao2024flashattention2,kwon2023efficient}.

Our main contributions are summarized as follows:
\begin{enumerate}[leftmargin=2em]
    \item We conduct a systematic empirical analysis verifying that the low-rank structure of visual tokens is consistent across models and datasets, with a highly stable dominant subspace.
    \item Based on this finding, we formulate visual token pruning as a lossy compression problem under a low-rank background, providing a principled surrogate objective that connects token selection with projection residuals onto the dominant subspace.
    \item We propose LRCP, which scores tokens by their projection residuals onto the dominant low-rank subspace. LRCP is training-free and compatible with efficient attention implementations, enabling plug-and-play integration into existing LVLM architectures.
    \item Extensive experiments on image and video benchmarks show that LRCP preserves over 94\% of the original performance with an 88.9\% token reduction, consistently outperforming existing methods under various compression ratios.
\end{enumerate}

\section{Related Work}

\subsection{Large Vision-Language Models}
Modern LVLMs couple a vision encoder with a language model via a lightweight projector, converting image patches into visual tokens for autoregressive decoding~\citep{liu2024visual,li2023blip2,dai2023instructblip}.
Within this paradigm, recent efforts have rapidly scaled the visual token budget: LLaVA-v1.5~\citep{liu2023improved} encodes a $336\times336$ image into 576 tokens, while LLaVA-NeXT~\citep{liu2024llavanext} raises this to several thousand via multi-crop encoding.
The same trend extends to native-resolution modeling~\citep{wang2024qwen2vl,bai2025qwen25vl,guo2025seed15vl}, multi-image understanding~\citep{zhu2025internvl3,li2024llavaonevision}, and long-video reasoning~\citep{lin2023videollava,zhang2025videollama3}.
Although longer visual sequences improve fine-grained perception, they also make visual tokens the dominant source of prefilling cost and inference latency, motivating effective token compression.
\subsection{Token Compression for LVLMs}
To reduce the inference cost introduced by visual tokens, recent studies have explored token compression for LVLMs. Existing methods can be grouped into attention-based~\citep{chen2024fastv,arif2025hired,zhang2024fastervlm,liu2025hiprune,yang2024visionzip} and representation-based~\citep{zhang2024sparsevlm,xing2024pyramiddrop,zhang2025vscan,ma2026apet,shang2024prumerge,ye2025atpllava,wang2025folder,xing2025cvc} approaches.
Attention-based methods, such as FastV~\citep{chen2024fastv}, estimate token importance from language-model attention scores and retain tokens relevant to the query or generation process.
Despite their effectiveness, their reliance on attention distributions can introduce positional bias and make them less compatible with efficient attention implementations~\citep{dao2022flashattention,dao2024flashattention2,shazeer2019fast}.
Representation-based methods instead reduce visual redundancy by exploiting relations among visual token features.
For instance, PyramidDrop~\citep{xing2024pyramiddrop} progressively drops tokens at successive LLM layers based on inter-token cosine similarity, removing those most similar to their neighbors.
These methods mainly focus on local token relations and rarely model the global representation structure of the full visual token set.
In contrast, our method studies token pruning from the perspective of low-rank compressibility, explicitly characterizes the dominant global structure of visual tokens, and avoids relying on attention weights.

\section{Method}

\subsection{Preliminaries: LVLMs}

Most current large vision-language models (LVLMs) follow a ViT-Projector-LLM architecture \citep{dosovitskiy2021vit,radford2021clip,vaswani2017attention,touvron2023llama}. Given an input image $\mathbf{I}\in\mathbb{R}^{H\times W\times 3}$ or a video $\mathbf{V}=\{\mathbf{v}_i\}_{i=1}^{T}$, the vision encoder first extracts visual features, yielding image-level embeddings $\mathbf{E}\in\mathbb{R}^{N\times D}$ or video-level embeddings $\mathbf{E}\in\mathbb{R}^{T\times N\times D}$. The projector then maps these visual embeddings into visual tokens aligned with the language space, denoted by $\mathbf{F}^{v}\in\mathbb{R}^{M\times D'}$ for images and $\mathbf{F}^{v}\in\mathbb{R}^{T\times M\times D'}$ for videos. These visual tokens are concatenated with text tokens $\mathbf{F}^{t}$ corresponding to the input instruction and passed into the LLM decoder. The decoder processes the multimodal context in a single prefilling pass, then autoregressively generates the response $\mathbf{Y}=\{y_j\}_{j=1}^{L}$. The conditional generation probability can be written as
\begin{equation}
p(\mathbf{Y}\mid \mathbf{F}^{v}, \mathbf{F}^{t})
=
\prod_{j=1}^{L}
p\left(y_j \mid \mathbf{F}^{v}, \mathbf{F}^{t}, \mathbf{Y}_{1:j-1}\right),
\end{equation}
where $\mathbf{Y}_{1:j-1}$ denotes the sequence of previously generated tokens before step $j$.

\subsection{Low-Rank Structure of Visual Tokens}

We characterize the low-rank structure of visual tokens by performing PCA \citep{jolliffe2002pca} on each layer's output across three models and three datasets. We measure effective dimensionality as the minimum rank required to reach a given fraction of explained variance:
\begin{equation}
\mathrm{Rank}_{l}@v
=
\min\left\{
d\in\mathbb{Z}^{+}:
\sum_{j=1}^{d}\lambda_l^j
\ge
\frac{v}{100}
\right\}.
\end{equation}
Here, $\lambda_l^j$ denotes the variance contribution of the $j$-th normalized principal component at layer $l$. We primarily report Rank@90\% and Rank@95\% as quantitative indicators. Since Qwen2.5-VL adopts a dynamic-resolution encoding mechanism, we fix its input visual token count to 1024 for consistent cross-model comparison.

\begin{figure*}[htbp]
\centering
\includegraphics[width=0.97\textwidth]{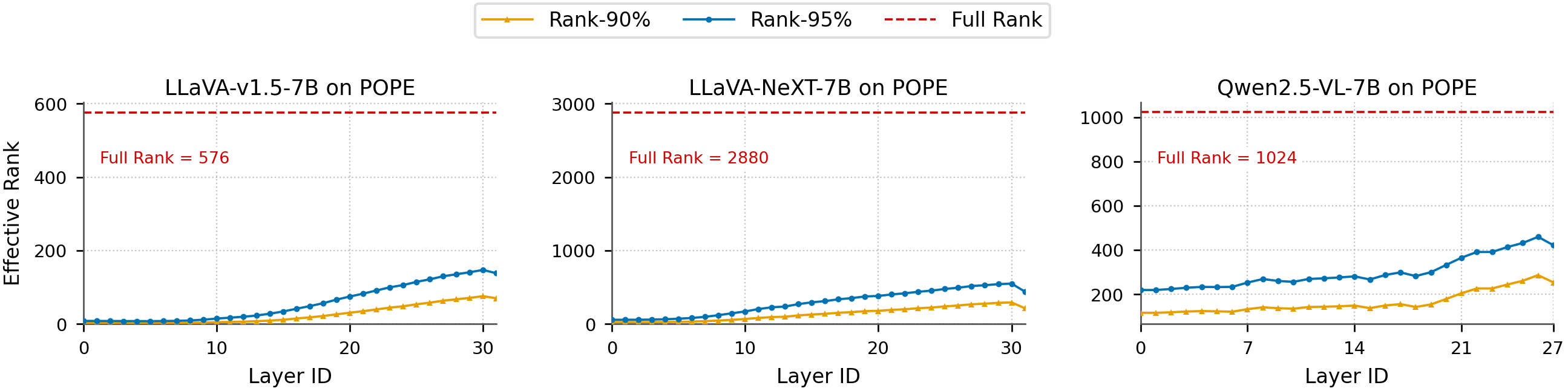}
\caption{Layer-wise Rank@90\% and Rank@95\% on POPE for LLaVA-v1.5-7B, LLaVA-NeXT-7B, and Qwen2.5-VL-7B. Lower values indicate stronger low-rank compressibility. Additional results are in Appendix~\ref{app:low_rank_additional}.}
\label{fig:rank_energy_stats_all}
\end{figure*}

Figure~\ref{fig:rank_energy_stats_all} provides cross-model Rank@90\% and Rank@95\% results on POPE. Across all three models, both indicators remain far below the full feature dimension, confirming that the low-rank structure is a consistent property rather than an isolated phenomenon. This implies that visual token redundancy is largely rooted in the shared low-rank structure, as most tokens concentrate near a common subspace and thus carry highly overlapping information. The plot also reveals a clear layer-wise trend: shallow layers are more low-rank, whereas deeper layers are more dispersed, likely because shallow layers primarily encode shared patterns while deeper layers capture finer semantic details.

\begin{figure*}[htbp]
\centering
\includegraphics[width=0.98\textwidth]{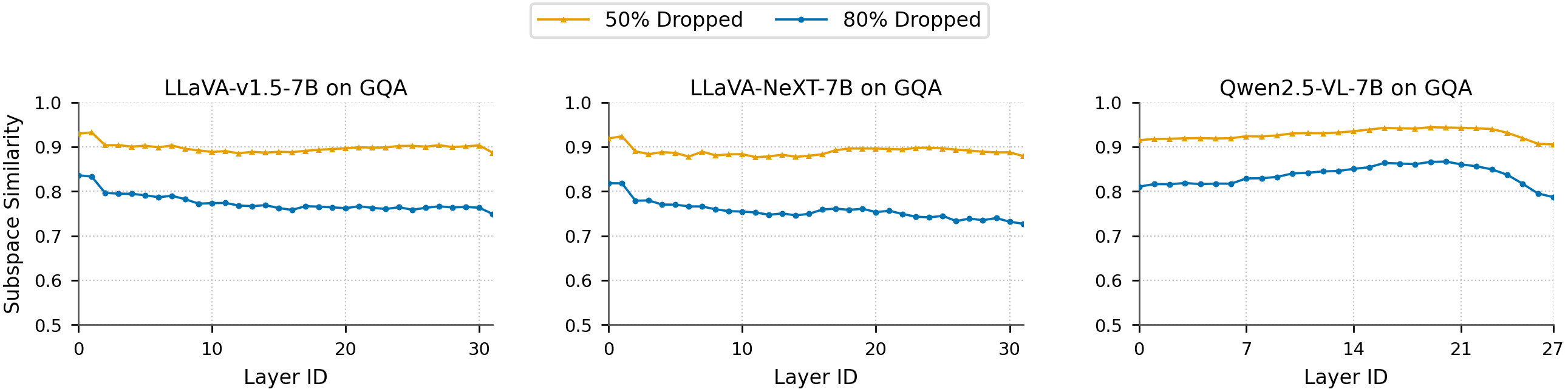}
\caption{Low-rank subspace stability on GQA for LLaVA-v1.5-7B, LLaVA-NeXT-7B, and Qwen2.5-VL-7B. High similarity after randomly dropping $50\%$ and $80\%$ of tokens confirms that the dominant subspace is a collective property. Additional results are in Appendix~\ref{app:low_rank_additional}.}
\label{fig:subspace_stability}
\end{figure*}

Furthermore, we examine the stability of the dominant subspace under token reduction by measuring the principal-angle-based similarity \citep{bjorck1973numerical} between the subspaces estimated from the full token set and from the remaining tokens after random dropout.
As shown in Figure~\ref{fig:subspace_stability}, subspace similarity remains high even after dropping $80\%$ of tokens across all three models, indicating that the dominant low-rank subspace originates from the shared structure of visual representations rather than depending on specific tokens. We further verify that this stability also holds under importance-based pruning (i.e., the actual token selection performed by LRCP), with detailed results provided in Appendix~\ref{app:low_rank_additional}.

Overall, these results suggest three key properties: (i)~visual tokens exhibit a pronounced low-rank structure that is consistent across models and datasets; (ii)~visual token redundancy is largely rooted in this shared low-rank structure; (iii)~the dominant low-rank subspace is highly stable under token reduction. These properties motivate us to explicitly leverage the global low-rank structure for visual token compression.

\subsection{A Low-Rank Compressibility Analysis}

Based on the empirical observations above, we view visual token compression as a lossy compression problem under a stable low-rank background. Given a visual token matrix \(X=[x_1,\dots,x_N]^\top \in \mathbb{R}^{N\times D}\), the goal is to select a subset $S$ with $|S|=K$ tokens to retain. The selection should ensure that the compressed representation preserves as much task-relevant information as possible. Formally, this objective can be written as
\[
S^\star
=
\arg\min_{S:\,|S|=K}
\mathcal{D}(X,\widehat{X}_S),
\]
where $\widehat{X}_S$ denotes the compressed representation induced by the retained tokens, and $\mathcal{D}(X,\widehat{X}_S)$ measures the distortion between the original visual representation and its compressed counterpart. In practice, the true task-level distortion is difficult to compute directly, since the contribution of each visual token depends on the multimodal projector, cross-token interactions in the LLM, and the specific downstream task. We therefore introduce a tractable surrogate objective based on squared reconstruction error to characterize the geometric information loss caused by token selection.

Let $P_r \in \mathbb{R}^{D \times D}$ denote the orthogonal projection matrix onto the $r$-dimensional dominant subspace estimated from $X$. We decompose $X$ into a low-rank component and a residual component:
\[
X = XP_r + X(I-P_r).
\]
Here, $XP_r$ captures the shared low-rank component, while $X(I-P_r)$ contains the residual variations. Let $M_S \in \mathbb{R}^{N\times N}$ denote the diagonal row mask associated with the retained set $S$:
\[
(M_S)_{ii} =
\begin{cases}
1, & i \in S,\\
0, & i \notin S.
\end{cases}
\]

The empirical analysis in Section~3.2 shows that the dominant subspace remains close to the original one even after a large fraction of tokens is randomly removed. This suggests that the dominant low-rank structure is not determined by a few individual tokens, but rather forms a shared background across the token set. Based on this observation, our surrogate analysis treats the shared low-rank component as approximately preserved after compression and focuses on characterizing the selection-dependent loss in the residual component. Specifically, we use the following surrogate representation:
\[
\widehat{X}_S(P_r) = XP_r + M_SX(I-P_r).
\]
The corresponding surrogate compression loss is
\[
\begin{aligned}
\mathcal{L}_{\mathrm{LR}}(S;P_r)
&= \|X-\widehat{X}_S(P_r)\|_F^2 \\
&= \|X-[XP_r+M_SX(I-P_r)]\|_F^2 \\
&= \|(I-M_S)X(I-P_r)\|_F^2 \\
&= \sum_{i\notin S}\|x_i(I-P_r)\|_2^2 .
\end{aligned}
\]
Thus, under this surrogate objective, the selection-dependent loss is exactly the sum of the squared projection residuals of the discarded tokens with respect to the dominant subspace. This gives a direct surrogate criterion for pruning: tokens with larger projection residuals should be retained first, since removing them would incur a larger surrogate compression loss.

\subsection{Low-Rank-Based Pruning Method}

\begin{figure*}[t!]
\centering
\includegraphics[width=\textwidth]{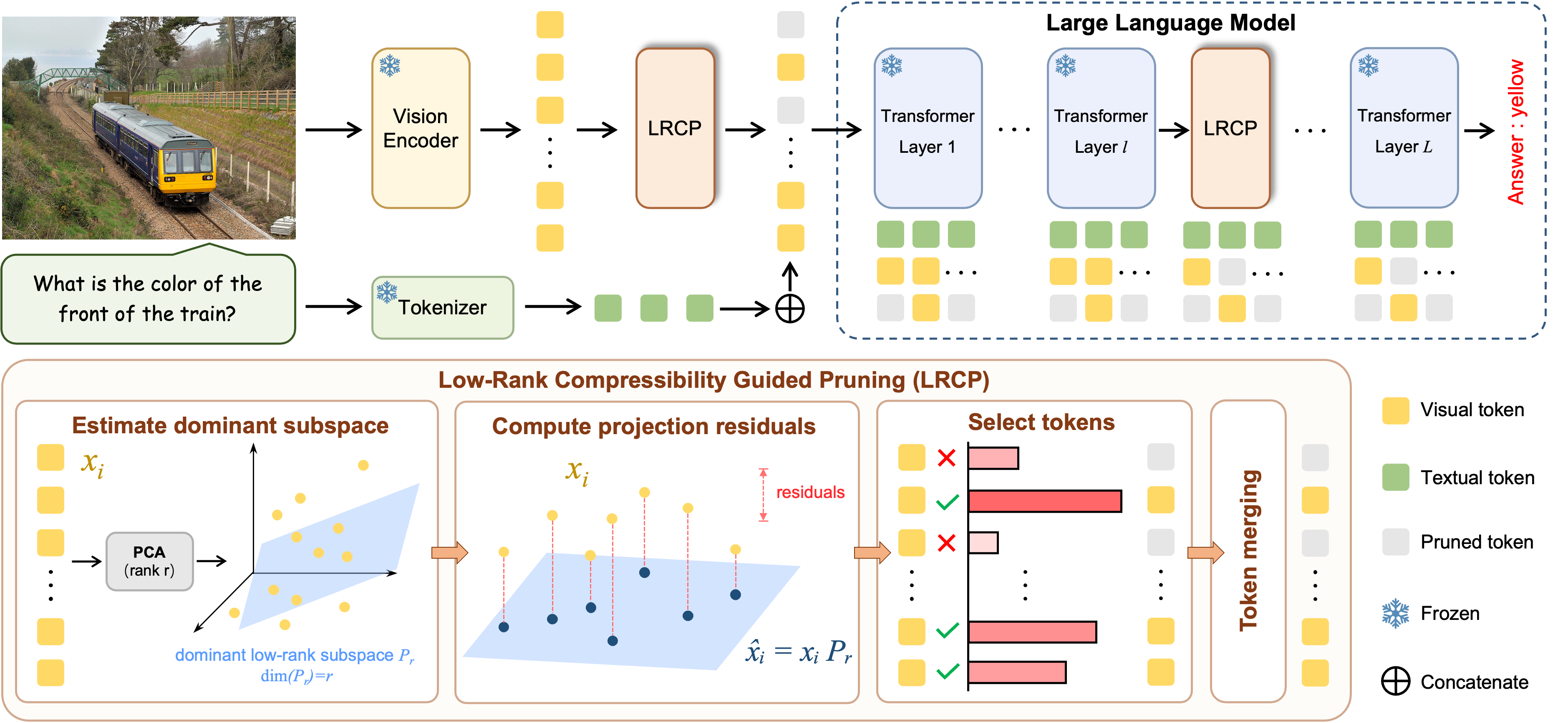}
\caption{Overview of LRCP. The method estimates the dominant low-rank subspace via PCA, retains tokens with large projection residuals, and merges discarded tokens into their nearest retained neighbors.}
\label{fig:method_overview}
\end{figure*}

Based on the above analysis, we use PCA to estimate the dominant low-rank subspace and score each token by its projection residual, which serves as a surrogate measure of token importance. We propose Low-Rank Compressibility Guided Pruning (LRCP). LRCP can be seamlessly inserted after the visual encoder output or into intermediate layers of the LLM. It does not rely on attention weights, and is therefore compatible with efficient attention implementations such as FlashAttention \citep{dao2022flashattention,dao2024flashattention2}. Specifically, LRCP consists of three stages: low-rank subspace construction, residual-based token ranking, and token merging. The complete pseudocode is provided in Algorithm~\ref{alg:lrcp} (Appendix~\ref{app:algorithm}).

\paragraph{Low-rank subspace construction.}

Given the token matrix output by the visual encoder, \(X=[x_1,\dots,x_N]^\top \in \mathbb{R}^{N\times D}\), and a retention budget \(K\), we set the PCA subspace dimension to \(r\). We first perform a low-rank PCA decomposition on the token matrix \citep{halko2011finding}, and take the top \(r\) principal components to span the dominant subspace. Let the basis of this subspace be
\begin{equation}
U_r \in \mathbb{R}^{D\times r},
\end{equation}
then the corresponding projection matrix is \(P_r=U_rU_r^\top\).

\paragraph{Projection residual computation.}

After obtaining \(U_r\), we compute the projection residual of each token with respect to the dominant subspace and use it as the importance score:
\begin{equation}
s_i=\|x_i(I-P_r)\|_2^2.
\end{equation}
Based on the projection residuals, we rank the visual tokens by importance and discard tokens with lower scores.

\paragraph{Token merging.}

To reduce the information loss caused by directly discarding tokens, we adopt a cosine-similarity-based token merging strategy inspired by prior work on token merging \citep{bolya2023tome}. For each token to be discarded, we assign it to the most similar retained token. Then, each retained token is updated by averaging it with all discarded tokens assigned to it, producing an updated feature representation.

\begin{table*}[t!]
\centering
\small
\setlength{\tabcolsep}{4pt}
\setlength{\belowcaptionskip}{4pt}
\renewcommand{\arraystretch}{1.08}
\caption{Performance comparison on 11 image understanding benchmarks with LLaVA-v1.5-7B (576 visual tokens). The last column reports the average performance retention relative to the uncompressed upper bound. The best result in each setting is in \textbf{bold} and the second-best is \second{underlined}.}
\label{tab:main_results}
\resizebox{\textwidth}{!}{
\begin{tabular}{lcccccccccccc}
\toprule
\textbf{Method} & \textbf{GQA} & \textbf{MMB} & \textbf{MMB$^{\mathrm{CN}}$} & \textbf{MME} & \textbf{POPE} & \textbf{SQA} & \textbf{VQA$^{\mathrm{V2}}$} & \textbf{VQA$^{\mathrm{Text}}$} & \textbf{SEED} & \textbf{MMVet} & \textbf{LLaVA-B} & \textbf{Avg.} \\
\midrule
\upperboundrow{13}{576 Tokens}
\midrule
{\color{flowtextgray}Vanilla} & {\color{flowtextgray}61.9} & {\color{flowtextgray}64.7} & {\color{flowtextgray}58.1} & {\color{flowtextgray}1862} & {\color{flowtextgray}85.9} & {\color{flowtextgray}69.5} & {\color{flowtextgray}78.5} & {\color{flowtextgray}58.2} & {\color{flowtextgray}59.3} & {\color{flowtextgray}31.1} & {\color{flowtextgray}66.9} & {\color{flowtextgray}100\%} \\
\midrule
\retainrow{13}{192}{66.7\%}
\midrule
FastV & 52.7 & 61.2 & 57.0 & 1612 & 64.8 & 67.3 & 67.1 & 52.5 & \second{57.1} & 27.7 & 49.4 & 88.3\% \\
SparseVLM & 57.6 & 62.5 & 53.7 & 1721 & 83.6 & \second{69.1} & 75.6 & 56.1 & 55.8 & \second{31.5} & 66.1 & 96.2\% \\
PDrop & 57.3 & 63.2 & 56.8 & 1766 & 82.3 & 69.0 & 75.1 & 56.1 & 54.7 & 30.5 & 65.8 & 96.2\% \\
VisionZip & 59.3 & 63.0 & 57.3 & 1778 & 85.2 & 68.7 & \second{76.6} & \second{57.3} & 56.4 & \best{31.7} & \best{67.7} & \second{98.1\%} \\
V2Drop & 58.5 & \best{63.7} & 56.6 & 1796 & 85.1 & \best{69.3} & 74.9 & 55.6 & 56.4 & 28.8 & 66.5 & 96.7\% \\
ApET & \second{60.2} & \second{63.4} & \best{57.9} & \second{1808} & \second{86.3} & 68.5 & 76.2 & 54.4 & 56.8 & 28.7 & 66.1 & 97.1\% \\
\rowcolor{flowgreenbg} \ours & \best{60.3} & \best{63.7} & \second{57.5} & \best{1823} & \best{86.4} & 68.8 & \best{76.9} & \best{57.4} & \best{57.2} & 31.2 & \second{67.5} & \best{98.8\%} \\
\midrule
\retainrow{13}{128}{77.8\%}
\midrule
FastV & 49.6 & 56.1 & 56.4 & 1490 & 59.6 & 60.2 & 61.8 & 50.6 & \best{55.9} & 28.1 & 52.0 & 84.4\% \\
SparseVLM & 56.0 & 60.0 & 51.1 & 1696 & 80.5 & 67.1 & 73.8 & 54.9 & 53.4 & 30.0 & 62.7 & 92.8\% \\
PDrop & 57.1 & 61.1 & \second{56.6} & 1644 & 82.3 & 68.4 & 72.9 & 54.8 & 53.3 & 30.8 & 61.9 & 94.1\% \\
VisionZip & \second{57.6} & 62.0 & 56.2 & 1759 & 83.2 & \best{68.9} & \second{75.6} & \best{56.8} & 54.9 & \best{32.1} & \best{64.8} & \second{96.6\%} \\
V2Drop & 56.3 & 61.8 & 54.5 & 1712 & 80.9 & \second{68.8} & 72.1 & 53.8 & 53.8 & 29.3 & 62.9 & 93.5\% \\
ApET & \best{58.9} & \second{62.3} & 56.4 & \second{1801} & \second{86.1} & 68.7 & 75.1 & 53.9 & 54.7 & 29.0 & \second{64.2} & 95.8\% \\
\rowcolor{flowgreenbg} \ours & \best{58.9} & \best{63.1} & \best{57.1} & \best{1807} & \best{87.0} & 68.5 & \best{75.8} & \best{56.8} & \second{55.4} & \second{31.8} & \best{64.8} & \best{97.7\%} \\
\midrule
\retainrow{13}{64}{88.9\%}
\midrule
FastV & 46.1 & 48.0 & 52.7 & 1256 & 48.0 & 51.1 & 55.0 & 47.8 & 51.9 & 25.8 & 46.1 & 75.3\% \\
SparseVLM & 52.7 & 56.2 & 46.1 & 1505 & 75.1 & 62.2 & 68.2 & 51.8 & 51.1 & 23.3 & 57.5 & 84.7\% \\
PDrop & 47.5 & 58.8 & 50.5 & 1092 & 55.9 & \best{69.2} & 69.2 & 45.9 & 40.0 & 30.7 & 59.2 & 81.8\% \\
VisionZip & 55.1 & 60.1 & \best{55.3} & 1687 & 77.0 & \second{69.0} & 72.4 & \second{55.5} & 52.2 & \best{31.5} & 62.9 & 93.4\% \\
V2Drop & 50.5 & 55.2 & 49.7 & 1470 & 75.1 & 68.9 & 71.2 & 51.8 & 51.4 & 29.1 & 62.4 & 88.3\% \\
ApET & \second{56.9} & \best{61.2} & 54.4 & \second{1714} & \second{84.4} & 68.9 & \second{72.5} & 53.0 & \best{52.4} & 29.4 & \second{63.0} & \second{93.6\%} \\
\rowcolor{flowgreenbg} \ours & \best{57.1} & \second{60.9} & \second{54.9} & \best{1721} & \best{85.3} & 68.5 & \best{73.4} & \best{55.6} & \second{52.2} & \second{31.3} & \best{63.1} & \best{94.7\%} \\
\bottomrule
\end{tabular}
}
\end{table*}

\section{Experiments}

\subsection{Experimental Setup}

\paragraph{Models and baselines.}
To evaluate the effectiveness of the proposed method, we conduct experiments on four representative vision-language models. For image understanding, we use LLaVA-v1.5-7B \citep{liu2023improved}, LLaVA-NeXT-7B \citep{liu2024llavanext}, and Qwen2.5-VL-7B \citep{bai2025qwen25vl}; for video understanding, we use Video-LLaVA-7B \citep{lin2023videollava}. We compare against FastV \citep{chen2024fastv}, SparseVLM \citep{zhang2024sparsevlm}, PDrop \citep{xing2024pyramiddrop}, VisionZip \citep{yang2024visionzip}, V2Drop \citep{chen2025v2drop}, and ApET \citep{ma2026apet}. The image benchmarks are GQA, MMBench, MMBench-CN, MME, POPE, ScienceQA, VQAv2, VQA$^{\mathrm{Text}}$, SEED, MMVet, and LLaVA-B \citep{hudson2019gqa,liu2024mmbench,fu2023mme,li2023pope,lu2022scienceqa,goyal2017vqa,singh2019textvqa,li2024seedbench,yu2023mmvet,liu2023improved}. The video benchmarks are TGIF \citep{jang2017tgif}, MSVD \citep{chen2011msvd}, and MSRVTT \citep{xu2016msrvtt}.

\paragraph{Implementation details.}
We follow the default inference configurations from the official implementations of all models. For image models, LRCP applies compression at both the output of the vision encoder and an intermediate LLM layer. Specifically, for the LLaVA family we apply intermediate-layer compression at layer 16, while for Qwen2.5-VL we apply it at layer 14. We provide the layer-wise ablation in Appendix~\ref{app:layer_ablation}. Qwen2.5-VL uses dynamic-resolution encoding and therefore produces a variable number of visual tokens per sample, so we fix the minimum and maximum number of pixels to 256 and 2048, respectively, for fair comparison across methods. For video tasks, Video-LLaVA encodes each video into 8 frames with 256 visual tokens per frame, yielding 2048 visual tokens in total. All experiments are conducted on NVIDIA A800 GPUs.

\subsection{Main Results}

\paragraph{Image understanding results.}
Table~\ref{tab:main_results} reports the main results on LLaVA-v1.5-7B. LRCP achieves the best average performance retention across different compression ratios, and the advantage widens as the token budget decreases. Under the most aggressive setting of retaining only 64 tokens, LRCP still preserves strong overall performance, exceeding the second-best method by 1.1\% in average performance retention. This suggests that explicitly modeling the global structure of the visual token set is more stable under extremely low token budgets.

\paragraph{High-resolution and dynamic-resolution results.}
LLaVA-NeXT-7B splits each input image into four local views and one rescaled global view, resulting in a much larger visual token budget than LLaVA-v1.5-7B. Table~\ref{tab:highres_results} shows that LRCP achieves the best overall results under all three token budgets of 640, 320, and 160, surpassing the second-best method by 0.7\%, 1.1\%, and 1.7\% in average performance retention, respectively. Table~\ref{tab:qwen25_results} shows a similar pattern on Qwen2.5-VL-7B. Under 20\% and 10\% average retention ratios, LRCP outperforms existing baselines by 1.2\% and 2.0\%, respectively, showing that the method adapts well to variable-length visual sequences.

\begin{table*}[t!]
\centering
\small
\renewcommand{\arraystretch}{1.18}
\begin{minipage}[t]{0.52\textwidth}
\centering
\setlength{\tabcolsep}{3pt}
\captionof{table}{Performance comparison on high-resolution image benchmarks with LLaVA-NeXT-7B (2880 visual tokens).}
\label{tab:highres_results}
\resizebox{\linewidth}{!}{
\begin{tabular}{lcccccccc}
\toprule
\textbf{Method} & \textbf{GQA} & \textbf{MMB} & \textbf{MMB$^{\mathrm{CN}}$} & \textbf{MME} & \textbf{POPE} & \textbf{VQA$^{\mathrm{V2}}$} & \textbf{VQA$^{\mathrm{Text}}$} & \textbf{Avg.} \\
\midrule
\upperboundrow{9}{2880 Tokens}
\midrule
{\color{flowtextgray}Vanilla} & {\color{flowtextgray}64.2} & {\color{flowtextgray}67.4} & {\color{flowtextgray}60.6} & {\color{flowtextgray}1851} & {\color{flowtextgray}86.5} & {\color{flowtextgray}81.8} & {\color{flowtextgray}61.3} & {\color{flowtextgray}100\%} \\
\midrule
\retainrow{9}{640}{77.8\%}
\midrule
SparseVLM & 60.3 & 65.8 & 58.5 & 1773 & 84.2 & 77.1 & 57.8 & 95.7\% \\
VisionZip & 61.3 & \second{66.2} & 59.1 & 1787 & 85.9 & 79.1 & \best{60.2} & 97.4\% \\
ApET & \second{63.0} & 65.3 & \second{59.3} & \second{1815} & \best{87.2} & \second{79.2} & 57.9 & \second{97.6\%} \\
\ours & \best{63.2} & \best{66.3} & \best{60.1} & \best{1827} & \second{86.3} & \best{79.6} & \second{58.9} & \best{98.3\%} \\
\midrule
\retainrow{9}{320}{88.9\%}
\midrule
SparseVLM & 59.3 & \best{64.2} & 55.9 & 1690 & 83.3 & 75.7 & \second{58.8} & 93.7\% \\
VisionZip & 59.3 & 63.1 & 56.3 & 1702 & 82.1 & \second{76.2} & \best{58.9} & 93.6\% \\
ApET & \second{61.0} & 63.5 & \second{56.6} & \best{1783} & \best{85.6} & 75.8 & 54.4 & \second{94.2\%} \\
\ours & \best{61.1} & \second{63.9} & \best{58.4} & \second{1759} & \second{84.8} & \best{76.7} & 57.4 & \best{95.3\%} \\
\midrule
\retainrow{9}{160}{94.4\%}
\midrule
SparseVLM & 51.2 & 52.1 & 48.6 & 1542 & 72.7 & 66.3 & 46.4 & 80.2\% \\
VisionZip & 55.5 & 60.1 & \best{55.1} & 1628 & 74.8 & 71.4 & \best{56.2} & 88.6\% \\
ApET & \second{58.4} & \second{60.8} & 52.3 & \second{1680} & \second{82.6} & \second{72.7} & 53.8 & \second{90.1\%} \\
\ours & \best{58.6} & \best{61.6} & \second{54.5} & \best{1693} & \best{83.3} & \best{74.6} & \second{55.8} & \best{91.8\%} \\
\bottomrule
\end{tabular}
}
\end{minipage}\hfill
\begin{minipage}[t]{0.46\textwidth}
\centering
\setlength{\tabcolsep}{6pt}
\renewcommand{\arraystretch}{1.235}
\captionof{table}{Performance comparison on dynamic-resolution image benchmarks with Qwen2.5-VL-7B.}
\label{tab:qwen25_results}
\resizebox{\linewidth}{!}{
\begin{tabular}{lcccccc}
\toprule
\textbf{Method} & \textbf{GQA} & \textbf{POPE} & \textbf{SQA} & \textbf{MME} & \textbf{MMB} & \textbf{Avg.} \\
\midrule
\upperboundrow{7}{256--2048 Tokens}
\midrule
{\color{flowtextgray}Vanilla} & {\color{flowtextgray}60.5} & {\color{flowtextgray}86.2} & {\color{flowtextgray}76.7} & {\color{flowtextgray}2327} & {\color{flowtextgray}83.3} & {\color{flowtextgray}100\%} \\
\midrule
\retainlabelrow{7}{Retain 20\% Tokens in Average}{80\%}
\midrule
SparseVLM & 54.7 & 73.6 & 71.6 & 2063 & 76.0 & 89.8\% \\
PDrop & 55.1 & 78.4 & 70.9 & 2117 & 77.3 & 91.6\% \\
VisionZip & 56.8 & 82.4 & \best{76.3} & 2134 & \best{79.3} & 95.2\% \\
V2Drop & 55.8 & 83.0 & 74.3 & 2196 & 77.6 & 94.6\% \\
ApET & \second{57.0} & \best{83.6} & 75.5 & \second{2211} & 77.4 & \second{95.5\%} \\
\ours & \best{57.6} & \second{83.5} & \second{76.0} & \best{2283} & \second{78.5} & \best{96.7\%} \\
\midrule
\retainlabelrow{7}{Retain 10\% Tokens in Average}{90\%}
\midrule
SparseVLM & 51.3 & 71.9 & 68.2 & 1849 & 71.7 & 84.5\% \\
PDrop & 52.0 & 74.8 & 69.7 & 1886 & 73.6 & 86.6\% \\
VisionZip & 52.4 & 78.9 & \second{74.1} & 2003 & \best{75.6} & 90.3\% \\
V2Drop & 52.5 & 78.7 & 73.6 & \second{2058} & 74.5 & 90.4\% \\
ApET & \second{53.4} & \best{79.3} & 74.0 & 2030 & 73.8 & \second{90.5\%} \\
\ours & \best{54.7} & \second{79.1} & \best{74.4} & \best{2168} & \second{74.9} & \best{92.5\%} \\
\bottomrule
\end{tabular}
}
\end{minipage}
\end{table*}

\paragraph{Video understanding results.}
Table~\ref{tab:video_results} summarizes the video understanding results on Video-LLaVA-7B under a unified budget of 256 retained tokens (an 87.5\% reduction from the original 2048 video tokens). LRCP achieves 97.8\% average accuracy retention and 99.3\% average score retention, surpassing the second-best method (VisionZip) by 3.4\% and 1.1\%, respectively. This demonstrates that LRCP generalizes well beyond image understanding to temporal visual sequences.

\begin{table*}[htbp]
\centering
\small
\renewcommand{\arraystretch}{1.15}
\begin{minipage}[t]{0.54\textwidth}
\centering
\setlength{\tabcolsep}{3pt}
\renewcommand{\arraystretch}{1.19}
\captionof{table}{Video understanding results on Video-LLaVA-7B (2048 visual tokens). All methods retain 256 tokens ($\downarrow$ 87.5\%).}
\label{tab:video_results}
\resizebox{\linewidth}{!}{
\begin{tabular}{lcccccccc}
\toprule
\multirow{2}{*}{\textbf{Method}} & \multicolumn{2}{c}{\textbf{TGIF}} & \multicolumn{2}{c}{\textbf{MSVD}} & \multicolumn{2}{c}{\textbf{MSRVTT}} & \multicolumn{2}{c}{\textbf{Avg.}} \\
 & Acc & Scr & Acc & Scr & Acc & Scr & Acc & Scr \\
\midrule
\multirow{2}{*}{{\color{flowtextgray}Vanilla}} & {\color{flowtextgray}46.9} & {\color{flowtextgray}3.34} & {\color{flowtextgray}69.8} & {\color{flowtextgray}3.91} & {\color{flowtextgray}57.1} & {\color{flowtextgray}3.49} & \multirow{2}{*}{{\color{flowtextgray}100\%}} & \multirow{2}{*}{{\color{flowtextgray}100\%}} \\
 & {\color{flowtextgray}100\%} & {\color{flowtextgray}100\%} & {\color{flowtextgray}100\%} & {\color{flowtextgray}100\%} & {\color{flowtextgray}100\%} & {\color{flowtextgray}100\%} & & \\
\midrule
\multirow{2}{*}{PDrop} & 40.3 & 3.21 & 61.5 & 3.74 & 41.8 & 3.19 & \multirow{2}{*}{82.4\%} & \multirow{2}{*}{94.4\%} \\
 & 85.9\% & 96.1\% & 88.1\% & 95.7\% & 73.2\% & 91.4\% & & \\
\multirow{2}{*}{VisionZip} & \second{44.3} & \second{3.29} & \second{65.2} & \second{3.83} & \second{54.5} & \second{3.43} & \multirow{2}{*}{\second{94.4\%}} & \multirow{2}{*}{\second{98.2\%}} \\
 & \second{94.5\%} & \second{98.5\%} & \second{93.4\%} & \second{97.9\%} & \second{95.4\%} & \second{98.3\%} & & \\
\multirow{2}{*}{\ours} & \best{46.5} & \best{3.34} & \best{67.7} & \best{3.87} & \best{55.6} & \best{3.45} & \multirow{2}{*}{\best{97.8\%}} & \multirow{2}{*}{\best{99.3\%}} \\
 & \best{99.1\%} & \best{100.0\%} & \best{97.0\%} & \best{99.0\%} & \best{97.4\%} & \best{98.9\%} & & \\
\bottomrule
\end{tabular}
}
\end{minipage}\hfill
\begin{minipage}[t]{0.44\textwidth}
\centering
\setlength{\tabcolsep}{3pt}
\captionof{table}{Ablation on subspace dimension $r$. Token retention ratios are 11.1\% for LLaVA-v1.5-7B and 10\% for Qwen2.5-VL-7B.}
\label{tab:sampling_ablation}
\resizebox{\linewidth}{!}{
\begin{tabular}{llcccc}
\toprule
\textbf{Model} & \textbf{Setting} & \textbf{GQA} & \textbf{POPE} & \textbf{SQA} & \textbf{MME} \\
\midrule
LLaVA-v1.5-7B & $r=4$ & \second{57.1} & \best{85.3} & \best{68.5} & \best{1721} \\
LLaVA-v1.5-7B & $r=8$ & \best{57.2} & \second{84.9} & \second{68.1} & \second{1696} \\
LLaVA-v1.5-7B & $r=12$ & 56.5 & 84.2 & 67.4 & 1692 \\
LLaVA-v1.5-7B & $r=16$ & 56.6 & 84.1 & 67.8 & 1694 \\
\midrule
Qwen2.5-VL-7B & $r=4$ & 54.1 & 77.1 & 74.2 & 2128 \\
Qwen2.5-VL-7B & $r=8$ & \best{54.7} & \best{79.1} & \second{74.4} & \best{2168} \\
Qwen2.5-VL-7B & $r=12$ & \second{54.2} & \second{77.5} & \best{74.5} & 2132 \\
Qwen2.5-VL-7B & $r=16$ & 53.9 & 77.1 & 74.1 & \second{2136} \\
\bottomrule
\end{tabular}
}
\end{minipage}
\end{table*}

\begin{table*}[htbp]
\centering
\begin{minipage}[t]{0.49\textwidth}
\centering
\small
\setlength{\tabcolsep}{5pt}
\renewcommand{\arraystretch}{1.27}
\captionof{table}{Ablation of token scoring criteria on LLaVA-v1.5-7B with 64 retained visual tokens. All variants use the same compression location.}
\label{tab:score_ablation}
\resizebox{\linewidth}{!}{
\begin{tabular}{lccccc}
\toprule
\textbf{Setting} & \textbf{GQA} & \textbf{POPE} & \textbf{SQA} & \textbf{MME} & \textbf{VQA$^{\mathrm{Text}}$} \\
\midrule
Low-rank projection norm & 56.1 & 81.0 & 67.4 & 1598 & 53.9 \\
Small residual first & 54.7 & 80.6 & 66.1 & 1532 & 54.2 \\
Residual only & \second{56.8} & \second{82.3} & \second{68.4} & \second{1612} & \second{55.3} \\
LRCP full & \best{57.1} & \best{85.3} & \best{68.5} & \best{1721} & \best{55.6} \\
\bottomrule
\end{tabular}
}
\end{minipage}\hfill
\begin{minipage}[t]{0.49\textwidth}
\centering
\small
\setlength{\tabcolsep}{5pt}
\renewcommand{\arraystretch}{1.15}
\captionof{table}{Ablation of dominant-subspace extraction strategies on LLaVA-v1.5-7B with 64 retained visual tokens.}
\label{tab:subspace_ablation}
\resizebox{\linewidth}{!}{
\begin{tabular}{lccccc}
\toprule
\textbf{Subspace} & \textbf{GQA} & \textbf{POPE} & \textbf{SQA} & \textbf{MME} & \textbf{VQA$^{\mathrm{Text}}$} \\
\midrule
Random directions & 51.1 & 82.5 & 64.2 & 1485 & 52.3 \\
Coordinate variance & 55.7 & \second{84.3} & 67.1 & 1579 & 53.9 \\
Cluster centers & \best{57.6} & 84.0 & \second{67.7} & \second{1706} & \second{54.8} \\
PCA & \second{57.1} & \best{85.3} & \best{68.5} & \best{1721} & \best{55.6} \\
\bottomrule
\end{tabular}
}
\end{minipage}
\end{table*}

\vspace*{-12pt} 
\subsection{Ablation Studies}

\paragraph{Effect of the subspace dimension $r$.}
The subspace dimension $r$ controls the boundary between shared structure and discriminative residuals. 
Table~\ref{tab:sampling_ablation} shows that LRCP remains stable across a broad range of $r$ values, although the best setting differs by model.
This is expected, as models with different degrees of low-rank compressibility require appropriately sized subspaces to capture the shared background.
We use $r=4$ for LLaVA-based models and $r=8$ for Qwen-based models. 
Furthermore, when $r$ becomes too large, the dominant subspace absorbs more fine-grained token variations, making the residual scores less discriminative.

\paragraph{Scoring criterion ablation.}
Table~\ref{tab:score_ablation} compares four scoring variants on LLaVA-v1.5-7B with 64 retained tokens: \textit{Low-rank projection norm} ($\| x_i P_r \|_2^2$ with merging, descending), \textit{Small residual first} ($\|x_i(I-P_r) \|_2^2$ with merging, ascending), \textit{Residual only} (projection-residual scoring without merging), and \textit{LRCP full} (projection-residual scoring with merging). 
The full LRCP variant consistently outperforms the alternatives, indicating that projection-residual-based scoring provides an effective measure of token importance and that token merging further improves performance.

\paragraph{Dominant-subspace extraction ablation.}
We fix the residual scoring rule and token budget, and compare four strategies for constructing the dominant subspace (Table~\ref{tab:subspace_ablation}): \textit{Random directions} (randomly sampled and orthogonalized directions), \textit{Coordinate variance} (the top-$r$ coordinate axes by variance), \textit{Cluster centers} (the subspace spanned by clustering centers), and \textit{PCA} (the top-$r$ principal components). PCA achieves the best overall performance, suggesting that constructing the subspace from the leading directions of variation yields residual scores that are more reliable for token selection.

\paragraph{Efficiency analysis.}
As shown in Table~\ref{tab:efficiency_analysis}, LRCP achieves the highest inference speedup on both LLaVA-v1.5-7B and LLaVA-NeXT-7B. On LLaVA-v1.5-7B, LRCP with 64 retained visual tokens achieves a $1.48\times$ overall speedup and a $1.59\times$ prefilling speedup. On LLaVA-NeXT-7B, the advantage further increases to $2.96\times$ and $3.00\times$, respectively.
LRCP also yields the lowest theoretical FLOPs in both settings. All timing results include the full overhead of each compression method; for LRCP this covers PCA computation, projection-residual scoring, token selection, and token merging.

\begin{table*}[htbp]
\centering
\small
\caption{Efficiency analysis on LLaVA-v1.5-7B (left) and LLaVA-NeXT-7B (right). We report total inference time (min:s), prefilling time, and TFLOPs. $\Delta$ denotes the speedup relative to the uncompressed baseline. All timings include the full overhead of each compression method and are measured on a single A800 GPU.}
\label{tab:efficiency_analysis}
\begin{minipage}[t]{0.49\textwidth}
\centering
\setlength{\tabcolsep}{4pt}
\renewcommand{\arraystretch}{1.15}
\resizebox{\linewidth}{!}{
\begin{tabular}{lcccccc}
\toprule
\multirow{2}{*}{\textbf{Methods}} & \multirow{2}{*}{\textbf{Token}} & \textbf{Total} & \multirow{2}{*}{\textbf{$\Delta\uparrow$}} & \textbf{Prefilling} & \multirow{2}{*}{\textbf{$\Delta\uparrow$}} & \multirow{2}{*}{\textbf{TFLOPs}} \\
 &  & \textbf{Time$\downarrow$} &  & \textbf{Time$\downarrow$} &  &  \\
\midrule
{\color{flowtextgray}LLaVA-v1.5-7B} & {\color{flowtextgray}576} & {\color{flowtextgray}17:05} & {\color{flowtextgray}1.00$\times$} & {\color{flowtextgray}102ms} & {\color{flowtextgray}1.00$\times$} & {\color{flowtextgray}8.82} \\
+ SparseVLM & 64 & 15:57 & 1.07$\times$ & 90.1ms & 1.13$\times$ & 2.31 \\
+ PDrop & 64 & 12:56 & 1.32$\times$ & 72.5ms & 1.41$\times$ & 2.16 \\
+ VisionZip & 64 & 12:10 & 1.40$\times$ & 69.2ms & 1.47$\times$ & 2.03 \\
+ V2Drop & 64 & 12:24 & 1.38$\times$ & 70.1ms & 1.46$\times$ & 2.12 \\
+ ApET & 64 & 11:46 & 1.45$\times$ & 65.2ms & 1.56$\times$ & 2.09 \\
\rowcolor{flowgreenbg} + LRCP & 64 & \best{11:31} & \best{1.48$\times$} & \best{64.0ms} & \best{1.59$\times$} & \best{2.02} \\
\bottomrule
\end{tabular}
}
\end{minipage}\hfill
\begin{minipage}[t]{0.49\textwidth}
\centering
\setlength{\tabcolsep}{4pt}
\renewcommand{\arraystretch}{1.15}
\resizebox{\linewidth}{!}{
\begin{tabular}{lcccccc}
\toprule
\multirow{2}{*}{\textbf{Methods}} & \multirow{2}{*}{\textbf{Token}} & \textbf{Total} & \multirow{2}{*}{\textbf{$\Delta\uparrow$}} & \textbf{Prefilling} & \multirow{2}{*}{\textbf{$\Delta\uparrow$}} & \multirow{2}{*}{\textbf{TFLOPs}} \\
 &  & \textbf{Time$\downarrow$} &  & \textbf{Time$\downarrow$} &  &  \\
\midrule
{\color{flowtextgray}LLaVA-NeXT-7B} & {\color{flowtextgray}2880} & {\color{flowtextgray}35:16} & {\color{flowtextgray}1.00$\times$} & {\color{flowtextgray}206ms} & {\color{flowtextgray}1.00$\times$} & {\color{flowtextgray}31.03} \\
+ SparseVLM & 160 & 30:26 & 1.16$\times$ & 136ms & 1.51$\times$ & 7.62 \\
+ PDrop & 160 & 13:45 & 2.56$\times$ & 79.5ms & 2.59$\times$ & 6.78 \\
+ VisionZip & 160 & 12:32 & 2.81$\times$ & 69.9ms & 2.95$\times$ & 4.72 \\
+ V2Drop & 160 & 12:49 & 2.75$\times$ & 73.5ms & 2.80$\times$ & 5.22 \\
+ ApET & 160 & 12:03 & 2.93$\times$ & 69.1ms & 2.98$\times$ & 4.74 \\
\rowcolor{flowgreenbg} + LRCP & 160 & \best{11:56} & \best{2.96$\times$} & \best{68.7ms} & \best{3.00$\times$} & \best{4.67} \\
\bottomrule
\end{tabular}
}
\end{minipage}
\end{table*}

\begin{figure*}[htbp]
\centering
\includegraphics[width=1.0\textwidth]{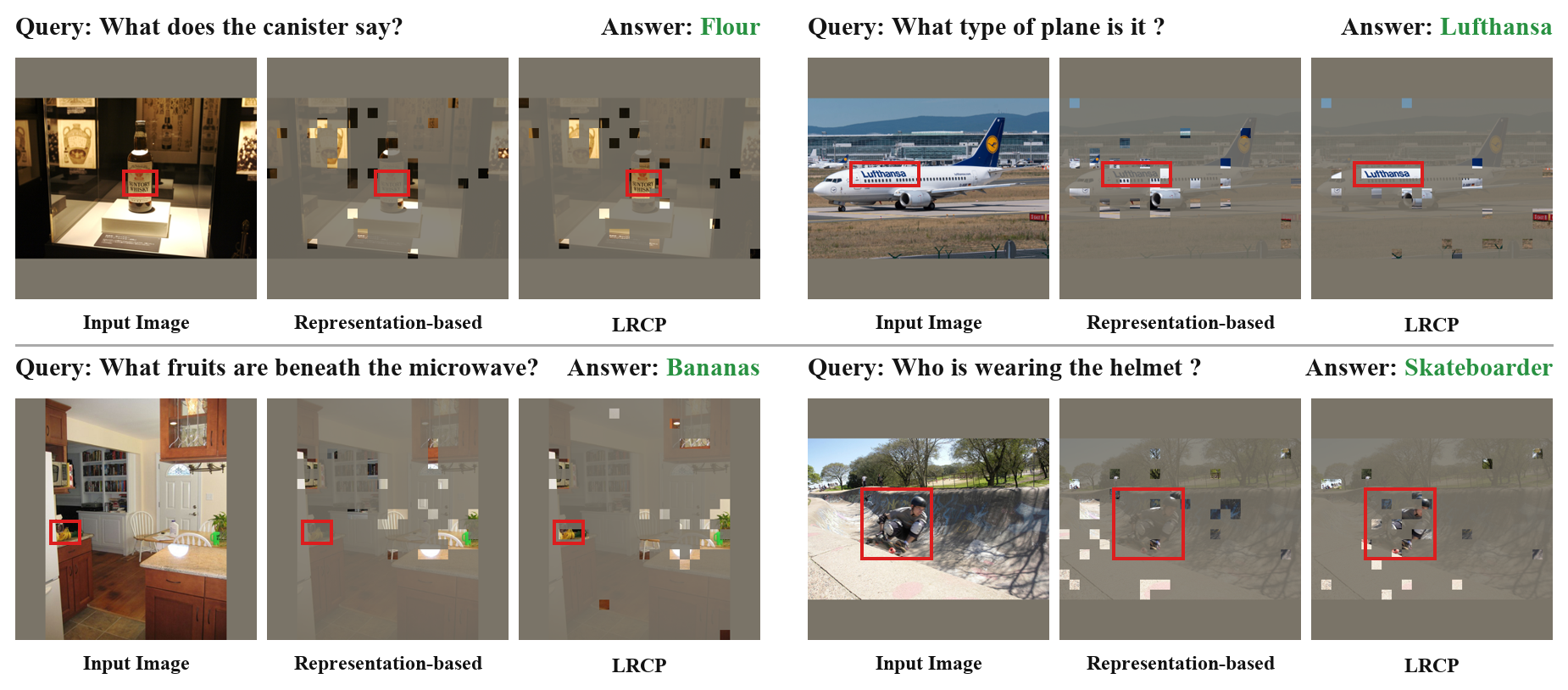}
\caption{Qualitative comparison of retained-token distributions on VQA$^{\mathrm{Text}}$ (top) and GQA (bottom). Retained tokens are highlighted; red boxes indicate answer-relevant regions. LRCP preserves more tokens in semantically informative areas.}
\label{fig:qualitative_results}
\vspace{-6pt}
\end{figure*}

\subsection{Qualitative Visualization}

To further illustrate the effectiveness of LRCP, we provide qualitative visualizations of token selection results in Figure~\ref{fig:qualitative_results}. We compare LRCP with a representation-based method, where we choose ApET~\citep{ma2026apet} as the baseline. LRCP more consistently retains text-bearing patches and fine-grained objects that are critical for answering the question, whereas the representation-based method tends to discard such discriminative tokens. These visualizations confirm that modeling the global structure of visual tokens enables more precise retention of irreplaceable visual information, leading to more robust token reduction.

\section{Conclusion}

We propose LRCP, a training-free visual token compression method guided by low-rank compressibility for efficient LVLM inference. 
Our analysis shows that visual tokens in LVLMs exhibit a stable low-rank structure across models, datasets, and layers, suggesting that visual token redundancy can be viewed as a consequence of shared low-rank representations. 
LRCP selects tokens according to their projection residuals with respect to the dominant low-rank subspace, aligning token selection with the global representation structure of visual tokens. 
Additionally, it is fully compatible with efficient attention implementations like FlashAttention.
Extensive experiments show that LRCP achieves a favorable balance between performance preservation and computational efficiency.

\section{Limitations}
This work has two limitations.
First, the optimal subspace dimension $r$ varies across model families and currently requires manual selection.
Second, evaluations are performed on public benchmarks, and additional validation in real-world application settings would strengthen the conclusions.

\section*{Acknowledgments}
This work is supported by the Natural Science Foundation of China under Grant No.~62305086, the China Postdoctoral Science Foundation under Grant No.~2023M740901, and the Natural Science Foundation of Heilongjiang Province of China under Grant No.~LH2024F032.

\bibliographystyle{unsrtnat}
\bibliography{references}

\newpage
\appendix

\section{Additional Experiments and Analysis}

\subsection{Additional Low-Rank Analysis on Other Datasets}
\label{app:low_rank_additional}

We extend the effective-dimensionality analysis to GQA and VQA$^{\mathrm{Text}}$, and the subspace stability analysis to POPE and VQA$^{\mathrm{Text}}$, verifying that the observed low-rank structure generalizes across datasets.

\begin{figure*}[htbp]
\centering
\includegraphics[width=0.97\textwidth]{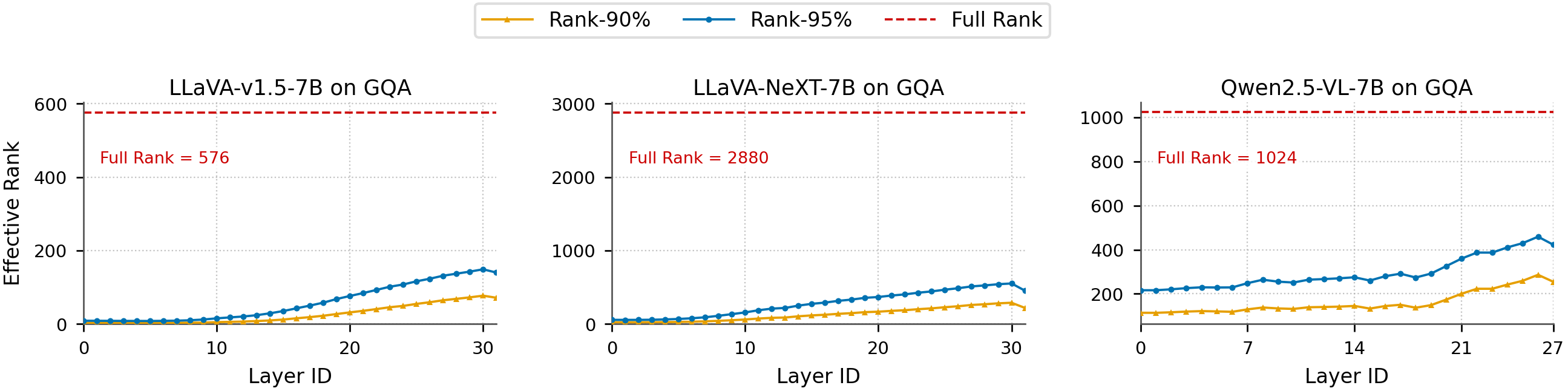}
\caption{Layer-wise Rank@90\% and Rank@95\% statistics on GQA for LLaVA-v1.5-7B, LLaVA-NeXT-7B, and Qwen2.5-VL-7B. The low-rank structure is consistent with the POPE results in the main text.}
\label{fig:rank_gqa_appendix}
\end{figure*}

\begin{figure*}[htbp]
\centering
\includegraphics[width=0.97\textwidth]{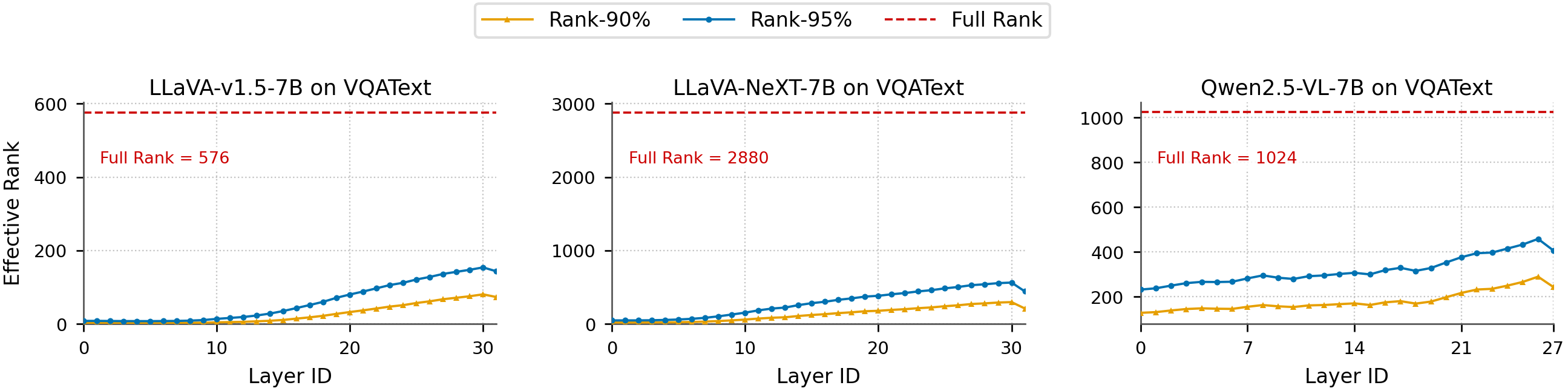}
\caption{Layer-wise Rank@90\% and Rank@95\% statistics on VQA$^{\mathrm{Text}}$ for LLaVA-v1.5-7B, LLaVA-NeXT-7B, and Qwen2.5-VL-7B. Despite the higher visual complexity of text-centric images, the low-rank property persists.}
\label{fig:rank_textvqa_appendix}
\end{figure*}

\paragraph{Effective dimensionality on additional datasets.}
Figures~\ref{fig:rank_gqa_appendix} and~\ref{fig:rank_textvqa_appendix} show the layer-wise Rank@90\% and Rank@95\% curves on GQA and VQA$^{\mathrm{Text}}$, respectively. The trends are highly consistent with the POPE results reported in the main text.
(i)~The effective rank remains substantially below the full feature dimension across all layers and models, reaffirming the pervasive low-rank compressibility of visual tokens.
(ii)~Shallow layers consistently exhibit lower effective rank than deeper layers, reflecting the transition from shared low-frequency visual patterns to more diverse, semantically rich representations.
(iii)~The cross-model ordering is preserved: LLaVA-v1.5-7B and LLaVA-NeXT-7B display similar rank profiles due to their shared ViT-based architecture, while Qwen2.5-VL-7B shows a slightly different trajectory owing to its dynamic-resolution encoding and distinct visual backbone.

Notably, VQA$^{\mathrm{Text}}$ images contain dense textual content and complex visual layouts, which might be expected to require higher-dimensional representations. Nevertheless, the rank statistics remain comparable, suggesting that the low-rank structure is an intrinsic property of how LVLMs encode visual information, rather than an artifact of image simplicity.

\begin{figure*}[htbp]
\centering
\includegraphics[width=0.98\textwidth]{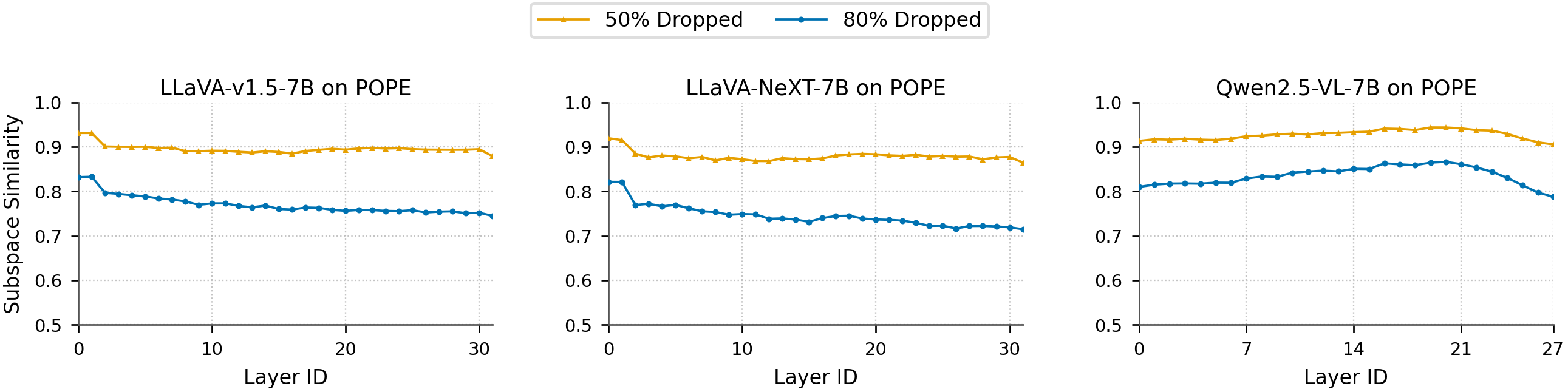}
\caption{Low-rank subspace stability on POPE for LLaVA-v1.5-7B, LLaVA-NeXT-7B, and Qwen2.5-VL-7B. Subspace similarity remains high even after removing 80\% of tokens.}
\label{fig:subspace_pope_appendix}
\end{figure*}

\begin{figure*}[htbp]
\centering
\includegraphics[width=0.98\textwidth]{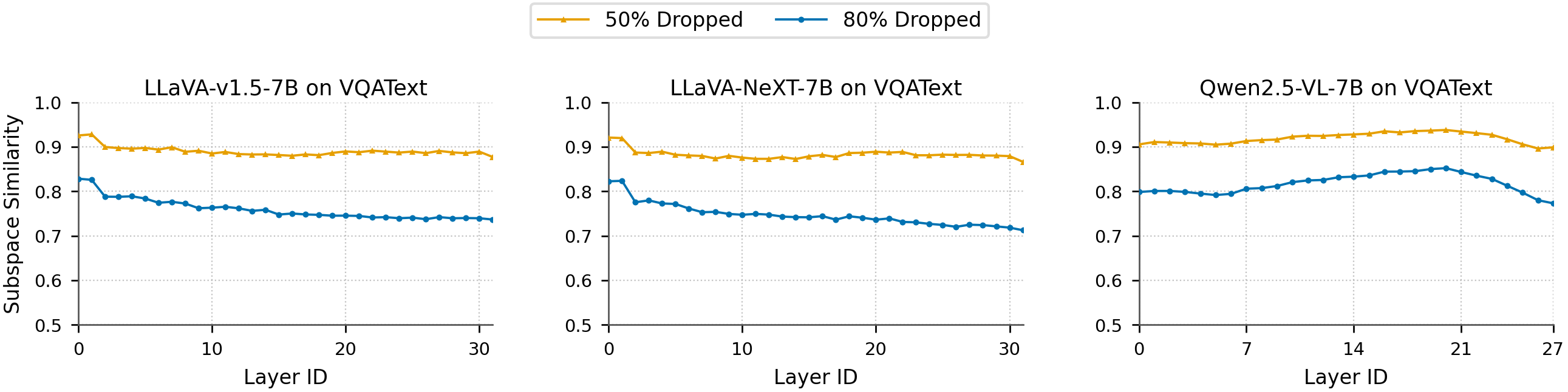}
\caption{Low-rank subspace stability on VQA$^{\mathrm{Text}}$ for LLaVA-v1.5-7B, LLaVA-NeXT-7B, and Qwen2.5-VL-7B. The stability is maintained across both text-dense and natural-scene images.}
\label{fig:subspace_textvqa_appendix}
\end{figure*}

\paragraph{Subspace stability on additional datasets.}
The main text reports subspace stability on GQA (Figure~\ref{fig:subspace_stability}). We now extend this analysis to POPE and VQA$^{\mathrm{Text}}$ (Figures~\ref{fig:subspace_pope_appendix} and~\ref{fig:subspace_textvqa_appendix}). Following the same protocol, we measure the principal-angle-based similarity~\citep{bjorck1973numerical} between the dominant subspace estimated from the full visual token set and the subspace re-estimated after randomly dropping 50\% and 80\% of the tokens. Across all models and both datasets, the subspace similarity remains consistently high (above 0.9 for most layers), even under the extreme 80\% dropout ratio. This corroborates the main-text finding that the dominant low-rank subspace arises from the collective structure of the visual token population rather than being anchored to a few specific tokens.

\paragraph{Subspace stability under importance-based pruning.}
The stability analyses above employ random token dropout as a stress test. Since LRCP preferentially retains tokens that \emph{deviate} from the dominant subspace, a natural follow-up question is whether the subspace remains stable under this non-random, importance-based selection. To investigate this, we compute the principal-angle-based subspace similarity between the full-set subspace and the subspace estimated from only the LRCP-retained tokens after each pruning stage.

\begin{figure*}[htbp]
\centering
\includegraphics[width=0.98\textwidth]{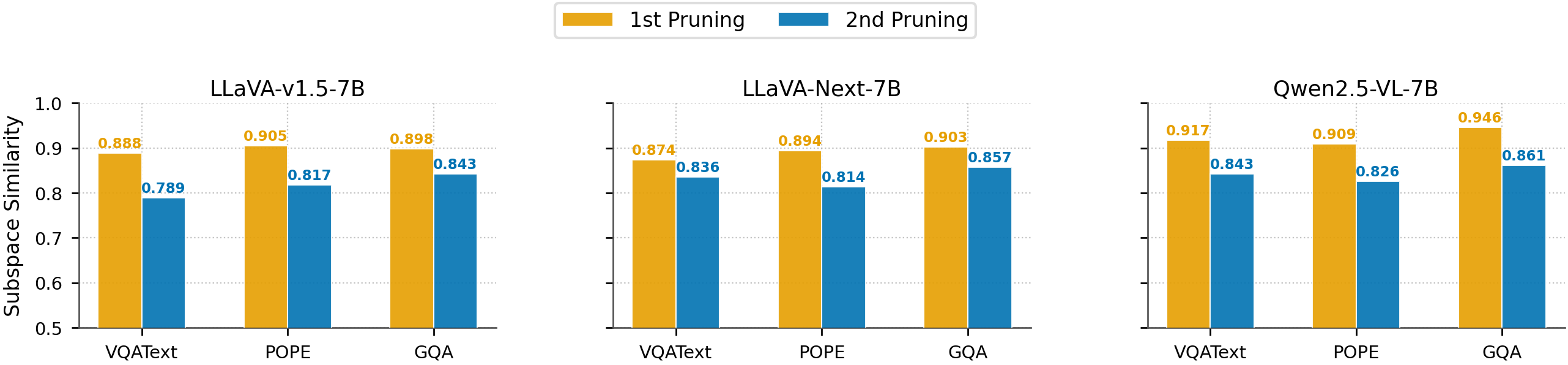}
\caption{Subspace stability under importance-based pruning by LRCP. We compare the dominant subspace estimated from the full token set with the subspace estimated from the retained tokens at each pruning stage. The consistently high similarity validates the assumption that LRCP preserves the low-rank structure.}
\label{fig:pruning_stages_stability}
\end{figure*}

As shown in Figure~\ref{fig:pruning_stages_stability}, the dominant subspace estimated from the retained tokens remains highly similar to the original across all measured stages. This observation can be explained as follows: tokens with large projection residuals still possess substantial components within the dominant subspace, which are sufficient to reconstruct the subspace with high fidelity. This stability under importance-based pruning provides a post-hoc justification for a key assumption in our surrogate analysis (Section~3.3), namely that the low-rank component is approximately preserved after compression.

\subsection{Algorithm Pseudocode}
\label{app:algorithm}

We provide the complete pseudocode of LRCP in Algorithm~\ref{alg:lrcp}. The procedure takes as input the visual token matrix $X\in\mathbb{R}^{N\times D}$, a retention budget $K$, and a subspace dimension $r$. It outputs the compressed token set $X'\in\mathbb{R}^{K\times D}$.

\begin{algorithm}[htbp]
\caption{Low-Rank Compressibility Guided Pruning (LRCP)}
\label{alg:lrcp}
\begin{algorithmic}[1]
\Require Visual tokens $X=[x_1,\ldots,x_N]^\top\in\mathbb{R}^{N\times D}$, token budget $K$, subspace rank $r$
\Ensure Compressed tokens $X'\in\mathbb{R}^{K\times D}$

\State Compute the top-$r$ PCA basis $U_r\in\mathbb{R}^{D\times r}$ of $X$
\State $s_i \leftarrow \|x_i(I - U_rU_r^\top)\|_2^2,\quad i=1,\ldots,N$
\State $\mathcal{S}\leftarrow \operatorname{TopK}_i(s_i,K)$, \quad $\mathcal{R}\leftarrow \{1,\ldots,N\}\setminus\mathcal{S}$
\For{$j\in\mathcal{S}$}
    \State $\mathcal{A}_j \leftarrow \{i\in\mathcal{R}: j=\arg\max_{k\in\mathcal{S}}\cos(x_i,x_k)\}$
    \State $x'_j \leftarrow \dfrac{x_j+\sum_{i\in\mathcal{A}_j}x_i}{1+|\mathcal{A}_j|}$
\EndFor
\State \Return $X'=[x'_j]_{j\in\mathcal{S}}^\top$
\end{algorithmic}
\end{algorithm}

\paragraph{Computational complexity.}
The dominant cost of LRCP lies in the truncated PCA decomposition (Line~1), which computes the top-$r$ principal components of $X$. Using a randomized SVD algorithm~\citep{halko2011finding}, this step requires $\mathcal{O}(NDr)$ operations, where $N$ is the number of visual tokens, $D$ is the feature dimension, and $r\ll \min(N,D)$ is the subspace rank. The projection residual computation (Line~2) costs $\mathcal{O}(NDr)$, and the top-$K$ selection (Line~3) takes $\mathcal{O}(N\log K)$. The cosine-similarity-based merging (Lines~4--7) involves computing pairwise similarities between the $(N-K)$ discarded tokens and $K$ retained tokens, costing $\mathcal{O}((N-K)KD)$ in the worst case. In practice, since $r$ is small (typically $r\le 8$) and the merging step operates on already-reduced token sets, the total overhead is negligible compared to a single LLM forward pass. We provide empirical latency measurements in the main text (Table~\ref{tab:efficiency_analysis}).

\subsection{Compression Layer Ablation}
\label{app:layer_ablation}

LRCP applies compression at two stages: the vision encoder output (fixed) and an intermediate LLM layer (a design choice; default: Layer~16 for LLaVA, Layer~14 for Qwen2.5-VL). Here we ablate the LLM compression layer while keeping the first stage unchanged. We sweep over layers \{8, 12, 14, 16, 18, 20\} for LLaVA-v1.5-7B (32 layers, 64 retained tokens) and \{8, 10, 12, 14, 16, 18\} for Qwen2.5-VL-7B (28 layers, 10\% average retention). Results are shown in Tables~\ref{tab:layer_ablation_llava} and~\ref{tab:layer_ablation_qwen}.

\begin{table}[htbp]
\centering
\small
\setlength{\tabcolsep}{5pt}
\renewcommand{\arraystretch}{1.15}
\caption{Ablation on the intermediate compression layer for LLaVA-v1.5-7B with an average of 64 retained visual tokens.}
\label{tab:layer_ablation_llava}
\begin{tabular}{lccccc}
\toprule
\textbf{Layer} & \textbf{GQA} & \textbf{POPE} & \textbf{SQA} & \textbf{MME} & \textbf{VQA$^{\mathrm{Text}}$} \\
\midrule
Layer 8  & 56.4 & 82.9 & 67.0 & 1619 & 55.3 \\
Layer 12 & \best{57.1} & 83.8 & 67.6 & 1667 & 55.4 \\
Layer 14 & \second{57.0} & 83.9 & 68.1 & 1702 & 55.4 \\
Layer 16 & \best{57.1} & \best{85.3} & \best{68.5} & \best{1721} & \second{55.6} \\
Layer 18 & 56.9 & 83.1 & \second{68.2} & 1625 & \best{55.8} \\
Layer 20 & 56.6 & \second{85.2} & 67.3 & \second{1709} & 55.0 \\
\bottomrule
\end{tabular}
\end{table}

\begin{table}[htbp]
\centering
\small
\setlength{\tabcolsep}{5pt}
\renewcommand{\arraystretch}{1.15}
\caption{Ablation on the intermediate compression layer for Qwen2.5-VL-7B with 10\% average token retention.}
\label{tab:layer_ablation_qwen}
\begin{tabular}{lccccc}
\toprule
\textbf{Layer} & \textbf{GQA} & \textbf{POPE} & \textbf{SQA} & \textbf{MME} & \textbf{MMB} \\
\midrule
Layer 8  & \second{54.4} & 78.3 & 74.1 & 2128 & 74.6 \\
Layer 10 & 54.1 & \second{80.1} & 73.8 & 2090 & 73.3 \\
Layer 12 & 54.1 & \best{81.0} & 73.7 & 2028 & 74.0 \\
Layer 14 & \best{54.7} & 79.1 & \best{74.4} & \best{2168} & \best{74.9} \\
Layer 16 & \second{54.4} & 78.1 & \second{74.3} & \second{2139} & \second{74.8} \\
Layer 18 & \second{54.4} & 78.3 & 74.1 & 2128 & 74.6 \\
\bottomrule
\end{tabular}
\end{table}

\paragraph{Analysis.}
Several observations emerge from Tables~\ref{tab:layer_ablation_llava} and~\ref{tab:layer_ablation_qwen}. First, for LLaVA-v1.5-7B, compressing at very early layers (e.g., Layer~8) consistently underperforms later layers across all benchmarks. This is expected: early LLM layers are still integrating cross-modal information from the visual encoder, and premature token removal disrupts this process. As the compression layer moves deeper, performance generally improves, with Layers~14--20 exhibiting comparable and near-optimal accuracy. We select Layer~16 as the default for LLaVA-based models, since it provides the best overall balance.

Second, for Qwen2.5-VL-7B, performance is notably more stable across compression layers, with narrow score ranges on all benchmarks. This robustness likely stems from Qwen2.5-VL's dynamic-resolution architecture, which may enable stronger cross-layer feature alignment. Layer~14 achieves the best results on most metrics and is therefore adopted as the default.

Overall, LRCP demonstrates low sensitivity to the compression layer choice in the middle-to-deep range, reducing the need for extensive hyperparameter tuning across architectures.

\subsection{Per-Stage Pruning Ratios}
\label{app:pruning_ratios}

The main text reports token budgets as the average number of visual tokens per layer. Table~\ref{tab:pruning_ratios} provides the detailed per-stage retention ratios for all experimental settings.

\begin{table}[htbp]
\centering
\small
\setlength{\tabcolsep}{6pt}
\renewcommand{\arraystretch}{1.15}
\caption{Per-stage retention ratios for each model. The compression layers are Layer~16 for LLaVA-based models (32 LLM layers) and Layer~14 for Qwen2.5-VL (28 LLM layers). ``Final Retain'' is the product of the two stages; ``Avg.\ Retain'' is the average retention ratio across all LLM layers, corresponding to the token budgets reported in the main text.}
\label{tab:pruning_ratios}
\begin{tabular}{llcccc}
\toprule
\textbf{Model} & \textbf{Setting} & \textbf{Stage 1 Retain} & \textbf{Stage 2 Retain} & \textbf{Final Retain} & \textbf{Avg. Retain} \\
\midrule
\multirow{3}{*}{LLaVA-v1.5-7B} & Avg 192 & 50.0\% & 33.3\% & 16.7\% & 33.3\% \\
 & Avg 128 & 33.3\% & 33.3\% & 11.1\% & 22.2\% \\
 & Avg 64  & 16.7\% & 33.3\% & 5.6\% & 11.1\% \\
\midrule
\multirow{3}{*}{LLaVA-NeXT-7B} & Avg 640 & 33.3\% & 33.3\% & 11.1\% & 22.2\% \\
 & Avg 320 & 16.7\% & 33.3\% & 5.6\% & 11.1\% \\
 & Avg 160 & 8.3\%  & 33.3\% & 2.8\% & 5.6\% \\
\midrule
\multirow{2}{*}{Qwen2.5-VL-7B} & Avg 20\% & 26.7\% & 50\% & 13.3\% & 20.0\% \\
 & Avg 10\% & 13.3\% & 50\% & 6.7\% & 10.0\% \\
\bottomrule
\end{tabular}
\end{table}

Across all models, the second-stage retention ratio is kept fixed (33.3\% for LLaVA-based models, 50\% for Qwen2.5-VL), while the first-stage ratio is adjusted to meet the target budget. This design concentrates the compression control in the first stage, keeping the second stage as a uniform refinement step.

\subsection{Supplementary Visualization}
\label{app:visualization}

To complement the qualitative analysis in the main text (Figure~\ref{fig:qualitative_results}), we provide additional visualization examples illustrating the token selection behavior of LRCP across different image types and compression levels.

\begin{figure*}[htbp]
\centering
\includegraphics[width=0.95\textwidth]{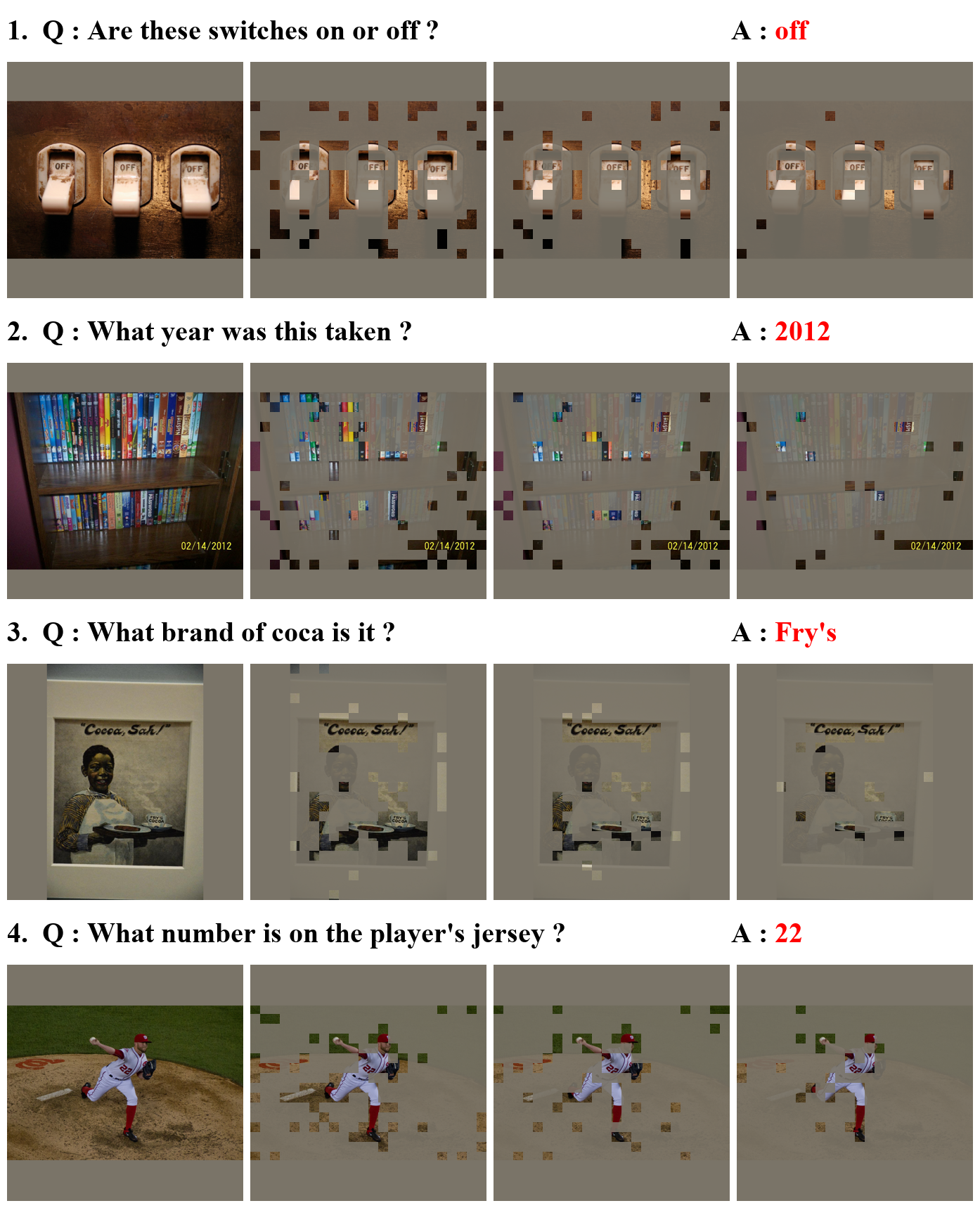}
\caption{Retained-token distributions of LRCP under three retention ratios. As the budget decreases, tokens in homogeneous backgrounds are progressively removed while semantically informative regions are preserved.}
\label{fig:qualitative_results_appendix}
\end{figure*}

Figure~\ref{fig:qualitative_results_appendix} shows LRCP's retained-token distributions under three different retention ratios. As the token budget decreases, LRCP progressively discards tokens in visually homogeneous regions while consistently preserving semantically salient areas such as object boundaries, fine-grained details, and regions relevant to the question. This is consistent with our theoretical motivation: background regions lie close to the dominant low-rank subspace and are compressed first, whereas discriminative regions with large projection residuals are preferentially retained.

\begin{figure*}[htbp]
\centering
\includegraphics[width=0.98\textwidth]{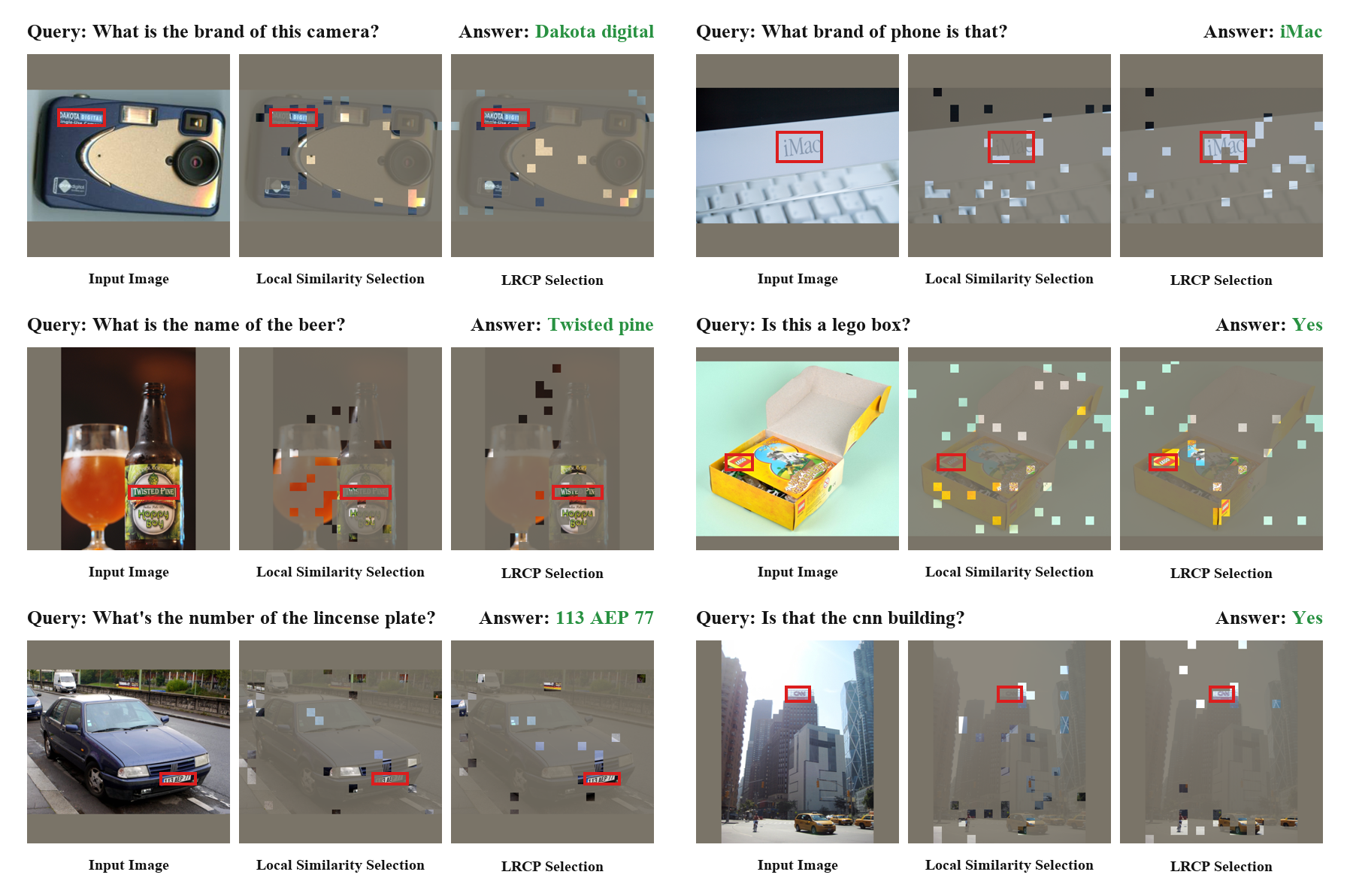}
\caption{Qualitative examples on VQA$^{\mathrm{Text}}$. For each case, we show the input image, the representation-based selection result, and the LRCP selection result. Red boxes indicate answer-relevant regions. LRCP more effectively preserves text-bearing regions.}
\label{fig:qualitative_results_appendix_textvqa}
\end{figure*}

Figure~\ref{fig:qualitative_results_appendix_textvqa} presents examples from VQA$^{\mathrm{Text}}$, where text-bearing regions occupy a small fraction of the image yet carry critical information. We compare LRCP with a representation-based baseline: the baseline spreads retained tokens across the image according to feature-level diversity, which may miss concentrated text regions. In contrast, LRCP assigns high importance scores to text regions because their fine-grained, high-frequency features deviate strongly from the low-rank visual background. As highlighted by the red boxes, LRCP more reliably retains tokens covering the relevant textual content, leading to more accurate answers.

\subsection{Broader Impact}
\label{app:broader_impact}

This work proposes a training-free visual token compression method for large vision-language models. We discuss both positive and potential negative societal implications below.

\paragraph{Positive impacts.}
By significantly reducing the number of visual tokens processed during inference, LRCP lowers the computational cost (FLOPs), memory footprint, and wall-clock latency of LVLM deployment. These efficiency gains have several positive downstream effects: (i)~\textit{Reduced energy consumption}: Large-scale LVLM serving is energy-intensive; token compression directly reduces the carbon footprint per inference call, contributing to more sustainable AI deployment. (ii)~\textit{Broader accessibility}: Lower resource requirements enable deployment on edge devices, mobile platforms, and in resource-constrained environments, democratizing access to multimodal AI capabilities. (iii)~\textit{Scalability}: As LVLMs continue to grow in parameter count and context length, efficient token management becomes critical for maintaining practical inference throughput in real-world applications.

\paragraph{Potential negative impacts.}
We do not foresee direct negative societal impacts from this work, as LRCP does not alter the pretrained model weights or introduce new training data; it only accelerates existing models at inference time. However, two indirect considerations merit discussion: (i)~As with any efficiency improvement, reduced deployment cost may lower the barrier to using LVLMs in applications that could raise ethical concerns (e.g., surveillance, content generation without consent). (ii)~Aggressive token pruning may inadvertently remove tokens encoding information about underrepresented visual content (e.g., small objects, fine-grained text, minority-class instances), potentially introducing or amplifying distributional biases in model outputs. We encourage practitioners to evaluate compression methods on diverse and representative benchmarks before deployment.

\end{document}